\documentclass[runningheads]{llncs}
\usepackage[T1]{fontenc}
\usepackage{graphicx}
\usepackage{float}
\usepackage[section]{placeins}
\usepackage{hyperref}
\usepackage{color}

\usepackage{esvect}
\let\oldvec\vec
\usepackage{amsmath}
\let\vec\oldvec
\usepackage{tabularx}
\usepackage{booktabs}

\begin{document}
\title{Link prediction Graph Neural Networks for structure recognition of Handwritten Mathematical Expressions}
\titlerunning{Link prediction GNNs for structure recognition of HMEs}
\author{Cuong Tuan Nguyen\inst{1}\orcidID{0000-0003-2556-9191} \and
Ngoc Tuan Nguyen\inst{1}\orcidID{0009-0002-8306-6269} \and
Triet Hoang Minh Dao\inst{1}\orcidID{0009-0001-1404-5926} \and 
Huy Minh Nhat Nguyen\inst{1}\orcidID{0009-0007-5539-1124} \and 
Huy Truong Dinh \inst{1}\orcidID{0000-0003-0290-6685}  
}
\authorrunning{C. T. Nguyen et al.}
\institute{Vietnamese-German University, Ho Chi Minh City, Vietnam 
\email{\{cuong.nt2,huy.td\}@vgu.edu.vn, \{21024003,10423179,10423045\}@student.vgu.edu.vn}\\
}
\maketitle              % typeset the header of the contribution
\begin{abstract}
We propose a Graph Neural Network (GNN)-based approach for Handwritten Mathematical Expression (HME) recognition by modeling HMEs as graphs, where nodes represent symbols and edges capture spatial dependencies. A deep BLSTM network is used for symbol segmentation, recognition, and spatial relation classification, forming an initial primitive graph. A 2D-CFG parser then generates all possible spatial relations, while the GNN-based link prediction model refines the structure by removing unnecessary connections, ultimately forming the Symbol Label Graph. Experimental results demonstrate the effectiveness of our approach, showing promising performance in HME structure recognition.

\end{abstract}
\section{Introduction}
Mathematical notations are essential in scientific documents for conveying concepts in mathematics and physics. With the emergence of digital pens and tablets, there is an increasing trend toward using handwritten mathematical notations as input, leading to a demand for systems that recognize these handwritten mathematical expressions (HMEs). We refer to the problem of online handwriting recognition, where each input sample is a sequence of trajectories captured by each pen-down called handwritten strokes.

Recognizing HMEs presents a challenging problem due to the complex two-dimensional structure of mathematical expressions. Early works focus on using structural parsing methods with a context-free grammar to represent mathematical expressions \cite{Yamamoto2006,lvaro2014,Le2016}, tree structure networks \cite{Zhang2020}, or graph-based algorithms \cite{Hu2016b}. However, due to the separated tasks of symbol segmentation, symbol recognition and structural analysis, these methods have limited overall expression recognition rate.

With the rise of deep learning algorithms, encoder-decoder approaches solve the problem of separated tasks by an end-to-end learning model. The encoder-decoder models developed from Recurrent Neural Networks (RNNs) decoder with attention \cite{Zhang2017,Zhang2019}, to the transformer decoder \cite{Zhao2021,Zhao2022} and multi-task learning \cite{Zhu2024}. These techniques focus on recognizing HME images and outputting one-dimensional character sequences in LaTeX format. However, they often fail to explicitly represent the hierarchical and spatial characteristics inherent to mathematical notations. Without clear symbol segmentation and two-dimensional relationships, these methods have limited effectiveness in practical scenarios that require structural clarity for computational manipulation and interactive editing.

Recent advancements have leveraged graph neural networks (GNNs) to address these limitations. Wu et al. \cite{Wu2021} proposed a graph-to-graph (G2G) learning method using stroke-level representations, modeling handwriting strokes as nodes and spatial relationships as edges. Xie et al. \cite{Xie2024} expanded this concept with an Edge Graph Attention Network (EGAT) focused on stroke-level interactions. Although stroke-level methods capture detailed structural information, they face the complexity of the graph due to numerous stroke primitives involved.

We propose a symbol-level graph recognition method for handwritten math expression recognition to overcome these limitations. By shifting from stroke-level to symbol-level representation, our approach reduces structural complexity of the graph. Our method integrates several key components:

First, we employ a Bidirectional Long Short-Term Memory (BLSTM) network for simultaneous symbol segmentation, recognition, and initial spatial relationship classification. Processing online handwritten data, this approach uses global contextual information to improve recognition of three tasks.

Next, we utilize a two-dimensional Cocke-Younger-Kasami (CYK) parsing algorithm to explore all possible spatial relationships among recognized symbols, constructing a symbol-level graph with recognized symbols as nodes and recognized relations as edges.

Finally, we further refine structural relationships using an Edge-featured Graph Attention Network (EGAT) on symbol-level graphs for a link prediction task. EGAT link prediction predicts whether an edge should be removed to produce the Symbol Layout Tree (SLT). 

In summary, the primary contributions of this paper are:
\begin{itemize}
    \item Formulate the problem of mathematical structure recognition to the problem of building symbol-level graph and link prediction.
    \item Propose a deep BLSTM and 2D CYK parser to construct symbol-level graph.
    \item Apply EGAT model for symbol-level link prediction to output mathematical expression as an SLT.
\end{itemize}

The remainder of this paper is structured as follows: Section 2 reviews related works, Section 3 describes our proposed methodology in detail, Section 4 presents experimental results and analysis, and Section 5 concludes with key insights and future research directions.

The source code for the experiments will be available in \\
\href{https://github.com/ntcuong2103/math_online_egat}{https://github.com/ntcuong2103/math\_online\_egat}

\section{Related works}

Handwritten Mathematical Expression (HME) recognition has evolved over recent decades, progressing from grammar-driven and heuristic-based methods to sophisticated deep learning and Graph Neural Network (GNN)-based approaches. The central challenge remains accurately parsing the complex, hierarchical two-dimensional structure inherent in mathematical expressions.

\subsection{Graph-Based Structural methods}

Hu and Zanibbi \cite{Hu2016b} formulate the problem of mathematical expression recognition as a Minimum Spanning Tree (MST) parsing by employing Edmonds' algorithm on directed Line-Of-Sight (LOS) graphs \cite{Hu2016a}. Their MST-based parser efficiently constructs expression trees by selecting edges that optimize spatial relationships among symbols, considerably reducing computational complexity and improving parsing accuracy compared to traditional CYK-based methods. Although MST approaches delivered promising improvements, their success heavily depended on accurate symbol segmentation and preliminary edge prediction.

Truong et al. used rule based split and merge methods for constructing the SLT \cite{Truong2022}. The method benefits from the global context of an HME for symbol and relation classification \cite{Nguyen2021}. 
\subsection{Graph Neural Network-Based methods}
Recent advancements have shifted towards deep learning-based models, particularly those leveraging Graph Neural Networks (GNNs), due to their intrinsic capability of modeling structured data. GNNs directly encode structural relationships within a graph representation, exploiting local context through iterative message-passing mechanisms to refine node and edge predictions.

\subsubsection{Stroke-Level GNN}
Stroke-level methods treat individual pen strokes as graph primitives, bypassing explicit symbol segmentation. Wu et al. \cite{Wu2021} introduced a pioneering Graph-to-Graph (G2G) model, conceptualizing HME recognition as graph transformation from raw stroke primitives to recognized symbol and relation graphs. Their approach employed sub-graph attention mechanisms to align input strokes precisely with their corresponding recognized symbols, simultaneously addressing segmentation and structural recognition tasks.

Building upon stroke-level concepts, Xie et al. \cite{Xie2024} proposed the Edge-weighted Graph Attention Network (EGAT) for HME structure recognition. EGAT integrates edge features into attention computations, directly predicting symbol identities and spatial relations from stroke graphs. Although stroke-level GNN methods are advantageous in reducing error propagation from segmentation stages, they inherently produce highly dense graphs, significantly increasing computational demands.

\subsubsection{Symbol-Level GNN}
Symbol-level GNN models address stroke-level complexity by pre-segmenting strokes into symbols. Tang et al. \cite{Tang2022} utilized a Graph Reasoning Network (GRN), structuring detected symbols from HME image into an SLT. GRN methods significantly enhance recognition accuracy by reducing the graph size and emphasizing meaningful symbol interactions, thus providing an efficient trade-off between segmentation complexity and graph representation fidelity.

\subsubsection{Summary}
Our method builds a symbol-level graph of recognized symbols and spatial relations between symbols, then applies GNN to predict whether an edge should be removed from the graph. The approach is similar to the MST, but we use GNN instead of Edmond's algorithm to remove edges. We also propose using the 2D CYK algorithm to build the symbol-level graph instead of the LOS algorithm.

\section{Methodology}
In this section, we propose a handwritten mathematical expression recognition method that constructs a symbol-level graph as input to the GNN for link prediction to refine the connections. 
\subsection{Overview of the method}
\begin{figure}
    \centering
    \includegraphics[width=1.0\linewidth]{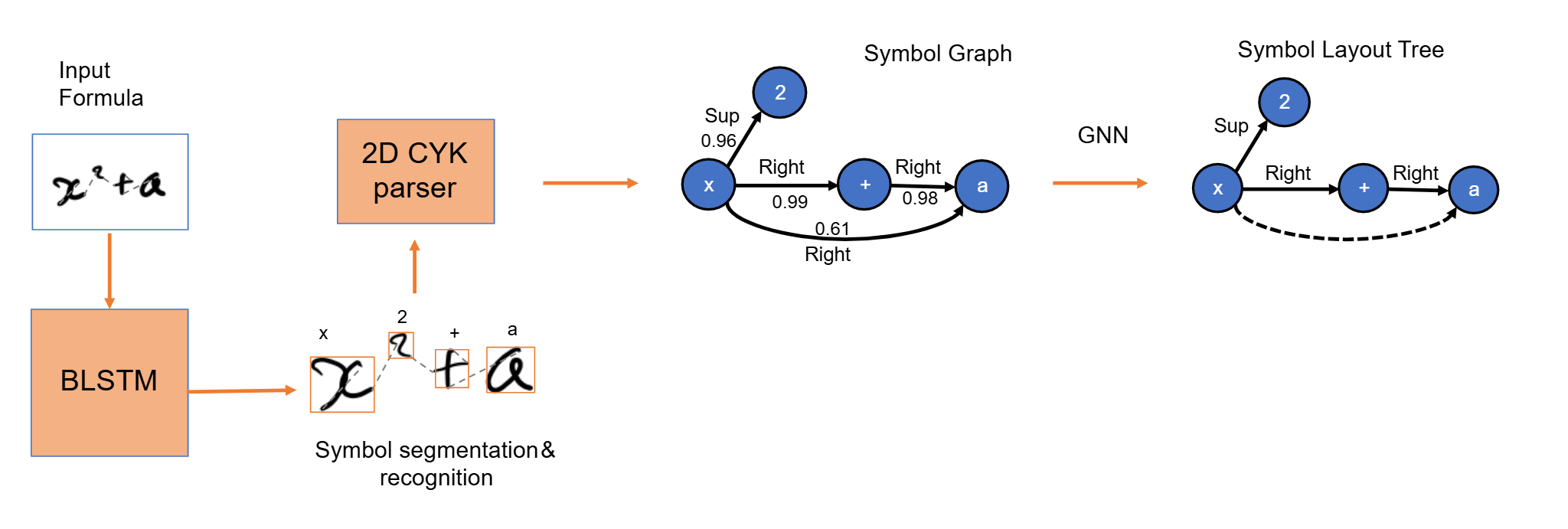}
    \caption{Overview of the method}
    \label{fig1}
\end{figure}

The pipeline of our method is shown in Fig. \ref{fig1}. A sequence of extracted features from an online HME input is fed to a deep BLSTM to predict symbols and relations. The segmented symbols are input to a symbol-level parser to build the symbol-level graph. Finally, the symbol-level graph is input to GNN for link prediction and produces SLT output. 

\subsection{Build symbol level graph}
This consists of two main stages as shown in Fig. \ref{fig2}: (1) symbol segmentation, recognition, and relation detection using a temporal recognition BLSTM network, and (2) symbol-level graph construction via a 2D CYK parsing algorithm.  The BLSTM outputs, at each time step corresponding to a recognized symbol or pair of symbols, a probability distribution over the relation classes. These probabilities seed our primitive graph construction.

\subsubsection{Symbol and Relation Recognition} 
This stage utilizes a deep bidirectional long short-term memory (BLSTM) network to process handwritten mathematical expressions. This model leverages the global context for symbol segmentation, symbol recognition, and relation classification \cite{Nguyen2021}. The method imposes a constraint to ensure the model outputs the relation classification at a precise time step, simultaneously addressing both symbol segmentation and relation classification. As shown in Level 0 of Fig. \ref{fig2}, the model takes the raw input sequence of strokes and segments them into symbols (e.g., 'x', '2', '+', 'a'). A key characteristic of this stage is that it primarily identifies relations between symbols that are consecutive in the writing order.
We currently restrict the BLSTM’s relation predictions to consecutive‐in-writing-order pairs to exploit temporal context. However, writing order can vary (e.g., users draw subscripts out of sequence). In practice this can cause missing edges for non-adjacent symbols. Thus, we resolve the problem by feeding the BLSTM arbitrary symbol pairs, which discussed in the next step of symbol-level graph construction.
\subsubsection{Symbol-Level Graph Construction} 
\begin{figure}[tb]
    \centering
    \includegraphics[width=1.0\linewidth]{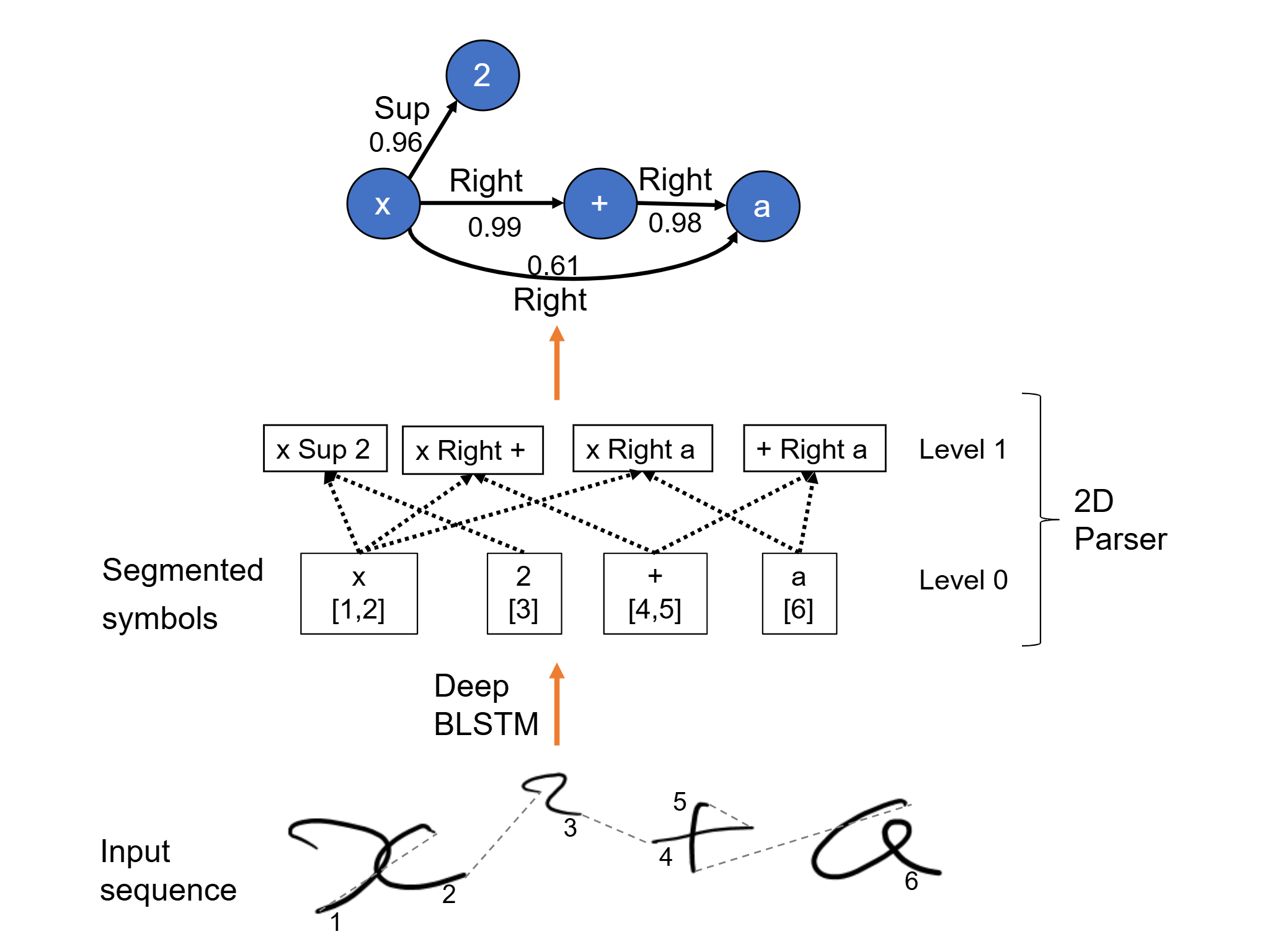}
    \caption{Symbol‐level graph constructed by deep BLSTM and 2D CYK parsing.}
    \label{fig2}
\end{figure}

The relations identified by the BLSTM are insufficient for capturing the full 2D structure. Therefore, in the second stage, we use a parser based on a 2D Stochastic Context-Free Grammar (2D-CFG) \cite{lvaro2014} and the Cocke-Younger-Kasami (CYK) algorithm to build a more complete primitive graph.
As depicted in Level 1 of Fig. \ref{fig2}, this parser considers all pairs of recognized symbols, not just consecutive ones. To determine the spatial relation between any two non-consecutive symbols (e.g., 'x' and 'a'), a new, temporary input sequence is constructed containing only the strokes for those two symbols. This new sequence is then fed into the same pre-trained BLSTM model to obtain a predicted spatial relation and its probability. This ensures all plausible pairwise relations are considered.

\subsubsection{Filter redundant edges using Line-Of-Sight}
We also applied the LOS algorithm \cite{Hu2016a} for filtering redundant edges. Let $E_\text{CYK}$ be the set of edges proposed by CYK algorithm, $E_\text{LOS}$ be the set of edges proposed by LOS algorithm. The output set of edges is:

\begin{equation}
    E_\text{CYK \& LOS} = E_\text{CYK} \land  E_\text{LOS}
\end{equation}
The result of this entire process is a dense, symbol-level graph, like the one shown in the "Symbol Graph" part of \ref{fig1}, which serves as the input to our GNN.

\subsection{Edge Graph Attention Networks for Link Prediction}
\begin{figure}
    \centering
    \includegraphics[width=1.0\linewidth]{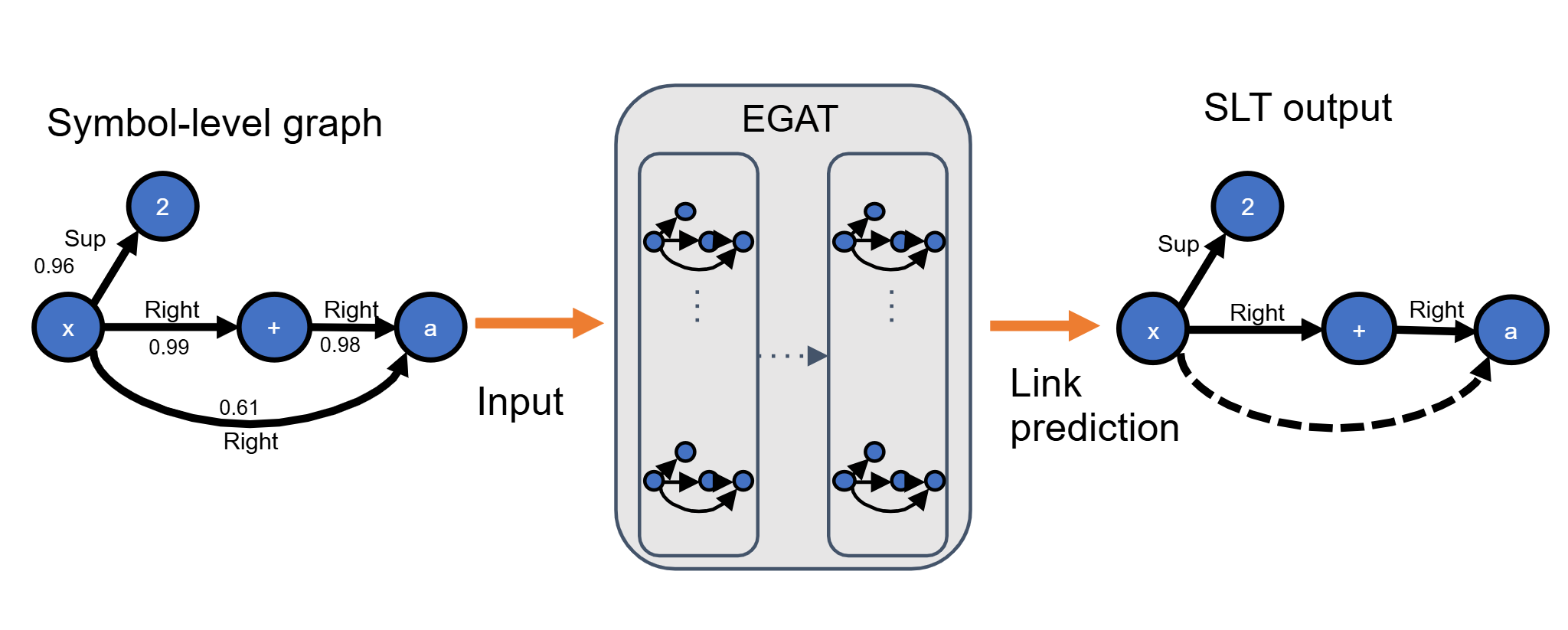}
    \caption{The link prediction EGAT refines the initial symbol‐level graph, removing edges to form an SLT.}
    \label{fig3}
\end{figure}

To further refine the symbol-level graphs generated by the CYK parsing and Line-of-Sight filtering, we utilize Edge Graph Attention Networks (EGAT). The EGATs are specifically designed to handle graph data by capturing rich structural relationships through attention mechanisms applied directly to edge features and node features. This capability makes EGAT particularly suitable for tasks involving intricate spatial relationships, such as handwritten mathematical expression recognition. Figure \ref{fig3} illustrates the EGAT that refines the initial symbol-level graph to form an SLT output.

\subsubsection{EGAT Architecture}
Our EGAT model is implemented using the Deep Graph Library (DGL) on a PyTorch backend. It comprises four stacked EGATConv layers \cite{Kamiski2022}, each of which updates node and edge embeddings via an edge-integrated attention mechanism. The final embeddings are passed to a two-layer MLP classification head, which outputs logits over the relation classes. All layers use LeakyReLU activations and 0.2 dropout to mitigate overfitting. The attention mechanism in EGAT is defined by:
\begin{equation}
    e_{ij} = F_\text{att} (f_{ij}^\prime)
\end{equation}
\begin{equation}
    f_{ij}^\prime = \text{LeakyReLU}(A [ h_i||f_{ij}||h_j])
\end{equation}
where $f_{ij}^\prime$ is the updated edge features, $h_i$ and $h_j$ represent the embedding vectors of nodes $i$ and $j$ respectively, $f_{ij}$ denotes the original edge features connecting nodes $i$ and $j$. $A$ is a learnable weight matrix that projects the concatenated features into a new embedding space suitable for attention computation. The operator $||$ indicates concatenation of two feature vectors.

\subsubsection{Input Features}
Our EGAT leverages both node and edge information:
\begin{enumerate}
    \item \textbf{Node Features} Each node’s feature is a one-hot vector encoding its recognized symbol class.
   \item \textbf{Edge Features } For each candidate edge, we take its upstream predicted relation (one-hot) and multiply it by the scalar confidence score output by the BLSTM. This produces a weighted edge embedding that encodes both the relation type and its confidence.
\end{enumerate}

\subsubsection{Link Prediction}
After propagation through the EGAT layers, a final MLP classifier predicts whether an edge should be included or not in the output of SLT.
The prediction for each edge uses a combined feature vector, which we term the NodeEdgeFeature. It is created by concatenating the final embeddings of the source node, the edge itself, and the destination node:
\begin{equation}
    \text{NodeEdgeFeature} = [ h_i||f_{ij}||h_j] 
\end{equation}
This combined feature is then passed to the MLP classifier. This ensures the final classification is conditioned on the full context of the edge and the symbols it connects. Instead of NodeEdgeFeature, the link can also be predicted from the EdgeFeature of $f_{ij}^\prime$.

\subsubsection{Training}
The loss of link prediction $L$ is computed using Binary Cross Entropy loss.  Training involves optimizing this loss across the entire symbol-level graph dataset, adjusting network parameters to minimize prediction errors. The Binary Cross-Entropy Loss:  

\begin{equation}
      L = \sum_{i=1}^N y_i \cdot \log p_i + (1 - y_i) \cdot \log (1 - p_i)
\end{equation}
where $y_i$ is the ground-truth label indicating whether an edge exists or not, and $p_i$ is the predicted probability, $N$ is the number of dataset samples.

 \section{Experiments}

\subsection{Datasets and metrics}

We evaluated our model using the CROHME (Competition on Recognition of Online Handwritten Mathematical Expressions) dataset. We used the train dataset from CROHME2019 and evaluated it on three test sets: CROHME2014, CROHME2016, and CROHME2019. The train dataset is used to train an LSTM model for symbol segmentation, symbol recognition, and relation classification, and used to train the EGAT model for link prediction.

For the deep BLSTM model, we used a stack of three BLSTM layers, each containing two LSTM layers (bidirectional) with 128 hidden units. The input sequence of coordiantes in a stroke was sampled by the Ramer method, then extracted four features: the sine and cosine of the writing direction, the normalized distance between two consecutive points and the pen-up/pen-down binary value. For the GNN, we use the stack of four EGAT layers followed by a fully connected layer for link prediction.

We measured the performance of symbol segmentation, symbol recognition, and relation classification using the BLSTM model. For symbol-level graph construction, we evaluated the coverage and redundancy of edges. 
Coverage measures the recall of ground-truth edges in the primitive graph:

\begin{equation}
     \text{Coverage} = \frac{|\{e | e \in G_\text{primitive} \land e \in G_\text{ground-truth}\}|}{|\{e | e \in G_\text{ground-truth}\}|}  \times 100\% 
\end{equation}

Redundancy measures the percentage of edges rather than that of ground-truth edges in the primitive graph:

\begin{equation}
    \text{Redundant} = \left( \frac{|\{e | e \in G_\text{primitive} \}|}{|\{e | e \in G_\text{ground-truth}\}|} - 1\right) \times 100\% 
\end{equation}

For link prediction, we evaluated link prediction accuracy. At the expression level, we evaluated the expression rate, the accuracy of recognizing entire mathematical expressions, and the structure rate, whether all the spatial relationships in a mathematical expression are correct.

\subsection{Experiments and results}

In the first evaluation, we evaluated the primitive graph produced by the LSTM model and the CYK parser. The result is shown in Table \ref{table_1}. We also evaluated the graph produced by the LOS algorithm and that by the CYK parser filtered by the LOS graph.

\begin{table}[ht]
\centering
\caption{Primitive graph evaluation on CROHME 2014 dataset}
\resizebox{\textwidth}{!}{%
\begin{tabular}{lccccc}
\toprule
\textbf{Method} & \textbf{Seg. (\%)} & \textbf{Seg. + Cls. (\%)} & \textbf{Rel. (\%)} & \textbf{Cov. (\%)} & \textbf{Red. (\%)} \\
\midrule
BLSTM-CYK & 98.09 (87.84) & 92.53 (60.37) & 92.99 (72.01) & 94.69 (79.16) & 107.65 \\
BLSTM-LOS & 98.09 (87.84) & 92.53 (60.37) & 92.64 (69.97) & 96.09 (84.78) & 722.30 \\
BLSTM CYK \& LOS & 98.09 (87.84) & 92.53 (60.37) & 92.64 (69.97) & 94.33 (76.92) & 87.74 \\
\bottomrule
\multicolumn{6}{l}{\footnotesize The value in parentheses is measured at expression level.} \\
\multicolumn{6}{l}{\footnotesize Abbreviation: Seg. (Segmentation), Cls. (Classification), Rel. (Relation)}, Cov. (Coverage), \\
\multicolumn{6}{l}{\footnotesize Red. (Redundant)} \\
\end{tabular}
}
\label{table_1}
\end{table}

Furthermore, the values in parentheses denote the evaluation metrics at the expression level. An expression is considered correctly recognized only if both its symbols and relations are accurately classified. As a result, these metrics set an upper bound on the overall expression-level performance.

Our analysis indicates that the LOS algorithm effectively recovers missing relationships but significantly elevates graph complexity due to redundant edges. Applying LOS as a filtering step with CYK reduces redundancy substantially with minimal sacrifice in coverage, suggesting an effective compromise between coverage and complexity.

Table \ref{table_2} presents the link prediction accuracies on the CROHME 2014, 2016, and 2019 test sets. The LOS algorithm achieves superior link prediction performance, with accuracies of 97.97\%, 97.99\%, and 98.12\% for each respective dataset. However, its high number of redundant edges adversely impacts the expression-level accuracy compared to alternative approaches. Notably, when CYK parsing is integrated with LOS filtering, there is a consistent improvement across all datasets with link accuracies of 92.61\%, 91.61\%, and 92.78\%, demonstrating the robust performance of this combined approach in balancing link prediction with overall expression-level reliability.

\begin{table}[ht]
\centering
\caption{EGAT link prediction evaluation}
\resizebox{0.8\textwidth}{!}{%
\begin{tabular}{lccc}
\toprule
 & \textbf{Test 2014} & \textbf{Test 2016} & \textbf{Test 2019} \\
\textbf{Input graph} & \textbf{Link acc. (\%)} & \textbf{Link acc. (\%)} & \textbf{Link acc. (\%)} \\
\midrule

BLSTM-CYK & 92.21 (57.00) & 91.49 (56.22) & 92.21 (53.97) \\
BLSTM-LOS & 97.97 (53.22) & 97.99 (52.67) & 98.12 (54.73) \\
BLSTM-CYK \& LOS & 92.61 (60.06) & 91.61 (57.09) & 92.78 (60.22) \\
\bottomrule
\multicolumn{4}{l}{\footnotesize The value in parentheses is measured at expression level} \\ \end{tabular}
}
\label{table_2}
\end{table}

Table \ref{table_3} illustrates the performance on full mathematical expression recognition across the three test sets. Our integrated CYK \& LOS approach achieved consistently higher Expression Rates (37.59\%, 38.35\%, 38.86\%) and Structure Recognition Rates (51.07\%, 48.60\%, 51.42\%) compared to individual CYK or LOS methods. These results confirm the effectiveness of combining structural parsing and redundancy filtering.

\begin{table}[ht]
\centering
\caption{Math expression evaluation}
\resizebox{\textwidth}{!}{%
\begin{tabular}{lcccccc}
\toprule
 & \multicolumn{2}{c}{\textbf{Test 2014}} & \multicolumn{2}{c}{\textbf{Test 2016}} & \multicolumn{2}{c}{\textbf{Test 2019}} \\
\textbf{Input graph} & \textbf{Exp. (\%)} & \textbf{Struct. (\%)} & \textbf{Exp. (\%)} & \textbf{Struct. (\%)} & \textbf{Exp. (\%)} & \textbf{Struct. (\%)} \\
\midrule
BLSTM-CYK & 36.36 & 49.23 & 38.18 & 48.42 & 36.07 & 47.28 \\
BLSTM-LOS & 34.83 & 47.09 & 37.01 & 46.11 & 37.66 & 48.62 \\
BLSTM-CYK \& LOS & 37.59 & 51.07 & 38.35 & 48.60 & 38.86 & 51.42 \\
\bottomrule
\end{tabular}
}
\label{table_3}
\end{table}

In the ablation study summarized in Table \ref{table_4}, we evaluated the impact of including NodeEdgeFeatures in our EGAT model. This inclusion notably improved link accuracy, validating the critical role of comprehensive feature representation in GNN-based parsing methods.

\begin{table}[ht]
\centering
\caption{Ablation study of EGAT network}
\resizebox{0.8\textwidth}{!}{%
\begin{tabular}{lccc}
\toprule
 & \textbf{Test 2014} & \textbf{Test 2016} & \textbf{Test 2019} \\
\textbf{Method} & \textbf{Link acc. (\%)} & \textbf{Link acc. (\%)} & \textbf{Link acc. (\%)} \\
\midrule
EGAT + EdgeFeature & 91.50 (55.16) & 91.11 (51.23) & 91.82 (55.11) \\
EGAT + NodeEdgeFeature & 92.61 (60.06) & 91.61 (57.09) & 92.78 (60.22) \\
\bottomrule
\end{tabular}
}
\label{table_4}
\end{table}

Finally, Table \ref{table_5} compares our GNN-based method with state-of-the-art models across multiple test sets. Although our approach does not surpass the top-performing systems such as MyScript or Samsung, it demonstrates competitive Expression rate and Structure Recognition Rates. Our method outperforms the graph-based method for building SLT such as the MST algorithm \cite{Hu2016b}, tree BLSTM \cite{Zhang2020} by a large margin. Compare to other GNN approaches which used GNN output features for symbol and relation classification, we are behind the expression rate since our method is limited by symbol classification performance at expression-level.

\begin{table}[ht]
\centering
\caption{Comparison to state-of-the-art models on CROHME datasets}
\resizebox{\textwidth}{!}{%
\begin{tabular}{llcccccc}
\toprule
 &  & \multicolumn{2}{c}{\textbf{Test 2014}} & \multicolumn{2}{c}{\textbf{Test 2016}} & \multicolumn{2}{c}{\textbf{Test 2019}} \\
\textbf{System} & \textbf{Type} & \textbf{Exp. (\%)} & \textbf{Struct. (\%)} & \textbf{Exp. (\%)} & \textbf{Struct. (\%)} & \textbf{Exp. (\%)} & \textbf{Struct. (\%)} \\
\midrule
MyScript & Gram & 62.68 & - & 67.65 & 88.14 & 79.15 & 90.66 \\
UPV/Wiris & Gram & 37.22 & - & 49.61 & 74.28 & 60.13 & 79.15 \\
Tokyo & Gram & 25.66 & - & 43.94 & 61.55 & 39.95 & 58.22 \\
Samsung & Gram & - & - & 65.76 & - & 79.82 & 89.32 \\

\midrule

MST & Graph & 26.88 & - & - & - & - & - \\
Tree-BLSTM& Graph & 29.91 & - & 27.03 & - & - & - \\
Tree-construct. & Graph & 44.12 & 58.62 & 41.76 & 49.43 & - & - \\

\midrule

TAP & EncDec & 46.90 & - & 44.80 & - & - & - \\
iFlyTek & EncDec & - & - & 57.02 & - & 80.73 & 91.49 \\

\midrule

G2G & GNN & - & - & 52.05 & - & - & - \\
LGM-EGAT & GNN & - & - & 56.41 & - & 58.22 & 83.07 \\
GGM-EGAT & GNN & - & - & 56.67 & - & 60.72 & 83.74 \\

\midrule

Our & GNN & 37.59 & 51.07 & 38.35 & 48.60 & 38.86 & 51.42 \\
\bottomrule
\end{tabular}
}
\label{table_5}
\end{table}

Our method demonstrates competitive performance, particularly when compared to other explicit graph-based methods like MST \cite{Hu2016b}, which it outperforms by a large margin. While our expression rates currently lag behind top-performing grammar-based (MyScript) and end-to-end encoder-decoder systems (iFlyTek), this highlights a key trade-off. Our pipeline approach separates symbol recognition from structural analysis, meaning errors from the initial BLSTM recognizer can propagate and limit the final expression rate.
However, the strength of our approach is its explicit structural output (a graph), which is more interpretable and editable than the LaTeX strings from encoder-decoder models. The strong Structure Rate (e.g., 51.42\% on CROHME 2019) shows that the GNN is highly effective at its specific task: learning and predicting the correct graph structure. This indicates significant potential, and future work focused on improving the initial symbol recognizer could close the gap with end-to-end systems. Furthermore, when comparing with other GNN-based methods like LGM-EGAT \cite{Xie2024}, our performance is comparable, indicating that the current state of symbol-level GNNs for this task is still evolving. The performance gap suggests that future research should focus on more robust primitive graph construction and more powerful GNN architectures.

\section{Conclusion}
In this paper, we presented a novel framework for handwritten mathematical expression recognition that effectively captures structural relationships using symbol-level graph representations. Our approach addresses key limitations of existing stroke-level graph recognition methods by reducing the graph complexity. Integrating a BLSTM network for simultaneous segmentation, recognition, and spatial relation classification provided robust accuracy on three tasks. The subsequent use of a CYK parsing algorithm, coupled with a practical Line-of-Sight heuristic, generated a clear and concise initial graph structure. Finally, our customized EGAT architecture accurately predicted and refined spatial relationships, producing Symbol Layout Tree as output. Experimental evaluations on the CROHME benchmark dataset demonstrated our proposed symbol-level graph approach's competitive performance in recognition accuracy.

Future research directions include exploring improvements on symbol-level graph construction by analyzing the failure cases, improving link prediction with graph data augmentation and further optimizing computational efficiency for real-time applications.

\bibliographystyle{splncs04}
\bibliography{icdar2025}

\end{document}